\documentclass[final]{cvpr}

\usepackage{times}
\usepackage{epsfig}
\usepackage{graphicx}
\usepackage{amsmath}
\usepackage{amssymb}
\usepackage{booktabs}
\usepackage{color}

\usepackage[utf8]{inputenc} 
\usepackage[T1]{fontenc}    
\usepackage{url}            
\usepackage{booktabs}       
\usepackage{amsfonts}       
\usepackage{nicefrac}       
\usepackage{microtype}      
\usepackage{multirow}
\usepackage{color}
\usepackage{mathtools}
\usepackage{siunitx}
\usepackage{graphicx}
\usepackage{grffile}
\usepackage{comment}

\usepackage{adjustbox}
\usepackage{caption}

\usepackage[pagebackref=true,breaklinks=true,colorlinks,bookmarks=false]{hyperref}

\def\confYear{CVPR 2021}

\def \short {} 
\ifx \short \undefined
	\newcommand{\cutsectionup}{}
	\newcommand{\cutsectiondown}{}

	\newcommand{\cutsubsectionup}{}
	\newcommand{\cutsubsectiondown}{}

	\newcommand{\cutparagraphup}{}

	\newcommand{\cuthalfcaptionup}{}
	\newcommand{\cuthalfcaptiondown}{}

	\newcommand{\cutcaptionup}{}
	\newcommand{\cutcaptiondown}{}

	\newcommand{\cuthalftablecaptionup}{}
	\newcommand{\cuthalftablecaptiondown}{}
\else
	\newcommand{\cutsectionup}{\vspace*{-2pt}}
	\newcommand{\cutsectiondown}{\vspace*{-2pt}}

	\newcommand{\cutsubsectionup}{\vspace*{-2pt}}
	\newcommand{\cutsubsectiondown}{\vspace*{-2pt}}

	\newcommand{\cutparagraphup}{\vspace*{-4pt}}

	\newcommand{\cuthalfcaptionup}{\vspace*{-12pt}}
	\newcommand{\cuthalfcaptiondown}{\vspace*{-10pt}}

	\newcommand{\cutcaptionup}{\vspace*{-12pt}}
	\newcommand{\cutcaptiondown}{\vspace*{-10pt}}

	\newcommand{\cuthalftablecaptionup}{\vspace*{-10pt}}
	\newcommand{\cuthalftablecaptiondown}{\vspace*{-10pt}}
\fi

\begin{document}

\title{S$^3$: Neural Shape, Skeleton, and Skinning Fields for 3D Human Modeling}

\author{
  Ze Yang$^{1,2}$\quad Shenlong Wang$^{1,2}$ \quad Sivabalan Manivasagam$^{1,2}$ \quad Zeng Huang$^{3}$ \\  \quad Wei-Chiu Ma$^{1,4}$ \quad Xinchen Yan$^{1}$\quad Ersin Yumer$^{1}$ \quad Raquel Urtasun$^{1,2}$\\ 
  \normalsize{$^{1}$Uber Advanced Technologies Group \quad $^{2}$University of Toronto, } \\
  \normalsize{$^{3}$University of Southern California \quad $^{4}$Massachusetts Institute of Techonology}\\
  \small\texttt{\{zeyang, slwang, manivasagam, urtasun\}@cs.toronto.edu},
  \\
  \small\texttt{zenghuan@usc.edu, weichium@mit.edu, \{xcyan, yumer\}@uber.com}
}

\twocolumn[{%
\renewcommand\twocolumn[1][]{#1}%
\maketitle
    \begin{center}
	\includegraphics[width=0.95\textwidth]{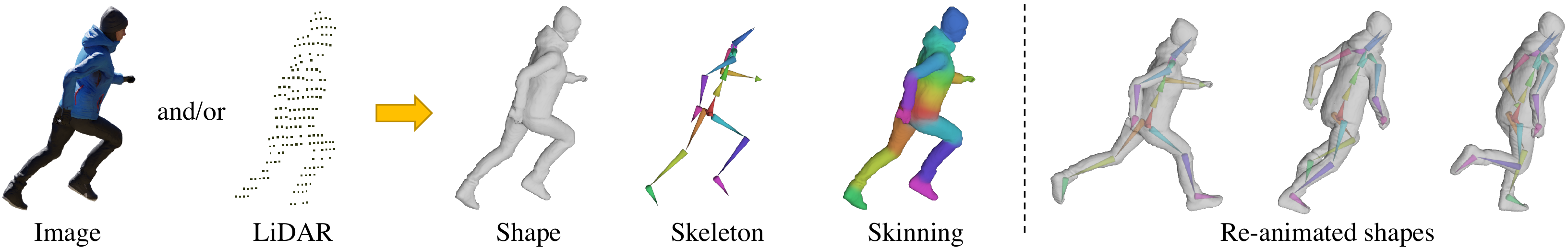}
    \end{center}
    \vspace{-5mm}
	\captionof{figure}{Given a single image and/or a single LiDAR sweep as input, our model infers shape, skeleton and skinning jointly, which can then be used to generate animated 3D characters in novel poses.}
    \label{fig:teaser}
	\vspace{3mm}
}]

\begin{abstract}
Constructing and animating humans is an important component for building virtual worlds in a wide variety of  applications such as virtual reality or robotics testing in simulation.
As there are exponentially many variations of humans with different shape, pose and clothing, it is critical to develop methods that can automatically reconstruct and animate humans at scale from real world data.
Towards this goal, we represent the pedestrian's shape, pose and skinning weights as neural implicit functions that are directly learned from data. This representation enables us to handle a wide variety of different pedestrian shapes and poses without explicitly fitting a human parametric body model, allowing us to handle a wider range of human geometries and topologies. 
We demonstrate the effectiveness of our approach on various datasets and show that our reconstructions outperform existing state-of-the-art methods.
Furthermore, our re-animation experiments show that we can generate 3D human animations at scale from a single RGB image (and/or an optional LiDAR sweep) as input. 

\end{abstract}

\cutsectionup
\section{Introduction}
\cutsectiondown
Realistic, articulated human simulation is an important task with a wide range applications.
It helps to bring characters to life in video games and movies~\cite{weng2019photo}, provides realistic AR/VR experiences in the context of sports \cite{zhu2020reconstructing, zhang2020vid2player} and social media~\cite{chen2018joint}, and is starting to play an important role in realistic simulations for testing robotic systems in both indoor ~\cite{sigal2010humaneva,varol2017learning,monszpart2019imapper} and outdoor environments~\cite{Dosovitskiy17,ze2020recovering}.
Traditionally, human reconstruction and animation is a time-consuming manual process. 
An artistic designer needs to create a scale-appropriate \textit{"joint estimation"} that specifies the human's degrees of freedom, \textit{"reconstruct"} the human 3D mesh geometry, and \textit{"skin"} the mesh by describing how the vertex positions deform as a function of the skeleton joints pose. 
Finally, the artist must specify the pose sequence that enacts an animation.
This manual approach is neither cost-effective nor efficient if we want to reconstruct and animate 3D humans at scale, for example by digitizing millions of different pedestrians observed in urban city scenes. 

Automating the human reconstruction and animation is very challenging as there are large variations in pedestrian shape, pose, clothing, and accoutrement.
Most existing animatable human modeling methods are pipelined systems consisting of a set of modules that run separately~\cite{vlasic2008articulated, stoll2010video, weng2019photo}. 
Typically, the first module performs joint estimation using marker~\cite{CMU_mocap} or markerless \cite{openpose} motion capture. 
From either dense 3D scans or images, the system then reconstructs shape~\cite{cheung2003shape, anguelov2005scape, loper2015smpl, kocabas2020vibe, DoubleFusion, alldieck2019tex2shape, kanazawa2018end, alldieck2019learning, he2020geo, saito2019pifu} and texture~\cite{tulsiani2017multi,  sun2018im2avatar, oechsle2019texture}.
The estimated pose skeleton is further associated with the mesh typically via closed-form skinning models \cite{DQB, kavan2005spherical} with hand-designed or automatic weight painting \cite{baran2007automatic, dionne2013geodesic}.
Unfortunately, for existing approaches to work successfully, they typically rely on an expensive 3D scanner in a controlled environment, a multi-camera cage \cite{vlasic2008articulated, stoll2010video}, or alternatively require relatively controlled viewpoints, typical frontal views with small variations \cite{weng2019photo}.
Few works so far have performed fully automatic end-to-end reconstruction and animation of clothed humans ~\cite{huang2020arch, xiang2020monoclothcap, bhatnagar2020combining}, but they all require fitting to a parametric body model, making it difficult to animate humans that deviate signficantly in geometry (i.e. non-tight clothing such as skirts and dresses). 

In this paper, we propose a scalable solution by reconstructing 3D animatable humans in the wild  that takes advantage of sensory data captured around our cities. 
However, this setting brings new challenges: in-the-wild data lacks ground-truth 3D shape and pose, making supervision for deep-learning based models challenging.
Furthermore, the captured sensory data can be noisy, of low-resolution, under non-canonical views and poses, and with various lighting conditions. These conditions are especially challenging for skinning, which requires accurate correspondence of each surface location to its corresponding body part. Inaccurate assignments will cause large aberrations during animation.
To address these challenges, we propose a novel approach that takes sensor data captured at a single viewpoint (i.e. image and/or LiDAR sweep) of a pedestrian and jointly predicts 3D mesh, skeleton joints, and skinning weights, all with a single network (see Figure~\ref{fig:teaser}).
The resulting animatable pedestrian can be directly deformed to novel poses and placed into new scenarios using either motion capture data or artist-created animations.
Inspired by the recent success of implicit modeling \cite{deepsdf, atzmon2020sal, chibane2020implicit} and neural radiance fields~\cite{mildenhall2020nerf}, we represent a 3D human as a continuous multi-dimensional neural field, which outputs the occupancy, human joint probability as well as skinning weights given each input 3D location in continuous space. 
This representation is very flexible and can capture fine details of clothed humans, handle different surface topologies (i.e. skirts), and adapt well to unseen human shapes as it is not constrained by a parametric model with a fixed (mesh) topology.
In addition, our end-to-end architecture overcomes the challenge of error propagation in conventional pipelines.

We demonstrate the effectiveness of our approach on both photorealistic synthetic human 3D data and a large-scale real-world  self-driving dataset.
Our approach  achieves better quantitative and qualitative performance compared to state-of-the-art methods in terms of shape reconstruction quality. 
Importantly, we also  show that we can  reliably re-animate the reconstructed 3D human given novel poses.

\cutsectionup
\section{Related work}
\cutsectiondown

\paragraph{Human Reconstruction:}
Early works such as SCAPE~\cite{anguelov2005scape} and SMPL~\cite{loper2015smpl} proposed to represent human body shape and pose-depedent variations using a parametric model.
Building upon them, there have been efforts in extending SMPL by modeling soft-tissue body motions~\cite{pons2015dyna}, integrating hand gestures~\cite{romero2017embodied} and facial expressions~\cite{joo2018total,pavlakos2019expressive}.
Later works proposed to estimate 3D pose and shape from a single image by minimizing the joint re-projection error~\cite{bogo2016keep, kanazawa2018end}, silhouettes re-projection error with a differentiable renderer~\cite{tung2017self, pavlakos2018learning}.
Recent works also explored the cross-frame consistency in monocular videos~\cite{kanazawa2019learning,pavlakos2019texturepose}, human-object interactions~\cite{zhang2020perceiving}, and part-specific attention~\cite{choutas2020monocular} to improve the reconstruction quality towards the next level.
One limitation of SMPL is that it only captures the shape space of naked human body, which inevitability leads to reality gap in practice.
Most recent works~\cite{zhu2019detailed,alldieck2019learning,ma2020learning,alldieck2019tex2shape} introduced per-vertex deformation to better reconstruct the clothed human body.

\cutparagraphup
\paragraph{Continuous Neural Representation:}
Recent years, deep implicit functions (or more generally, continuous neural field) has enjoyed increasing popularity
in 3D representation for static scenes or rigid objects~\cite{deepsdf,OccupancyNet,DISN} and deformable objects such as human body shapes~\cite{chibane2020implicit, saito2019pifu, saito2020pifuhd, he2020geo, zheng2020pamir}.
Those approaches generally have more degrees of freedom to capture the variations in human shape and clothing.
Continuous neural representation has also shown to achieve decent performance in novel view synthesis~\cite{mildenhall2020nerf,sitzmann2019srns} and object recognition~\cite{kirillov2020pointrend,yang2019dense}.

\cutparagraphup
\paragraph{Animating Human:}
A common approach in computer graphics to animate a human character is through skeletal animation. It requires an articulated object represented with two components: a surface representation and a hierarchical structure called skeleton. Skinning binds each surface vertex to the bones with skinning weights, measuring how the mesh's vertex will change when a bone moves.
Linear blend skinning (LBS)~\cite{lewis2000pose} linearly combines the transformation from each bone through skinning weights and applies the composed transformation to mesh vertices.
SMPL~\cite{loper2015smpl} extends LBS with identity-driven shape variations and pose-dependent shape variations.
Jointly reconstructing and animating humans from an image is an emerging domain in vision and graphics.
Recent approaches~\cite{weng2019photo,kanazawa2018end,kolotouros2019learning} reconstruct and animate from a single photo by using the SMPL~\cite{loper2015smpl} model.
The implicit function-based methods achieve better reconstruction quality in handling cloth and accessories. However, animation requires additional post-processing, such as generating a warp field guided by parametric model~\cite{huang2020arch}, or explicitly fitting the SMPL model as a post-processing step~\cite{bhatnagar2020combining}.
Unlike these work, we predict all the necessary components for animation: shape, skeleton, and skinning through a
single unified network.

\begin{figure*}[h]
\begin{center}
	\includegraphics[width=1\textwidth]{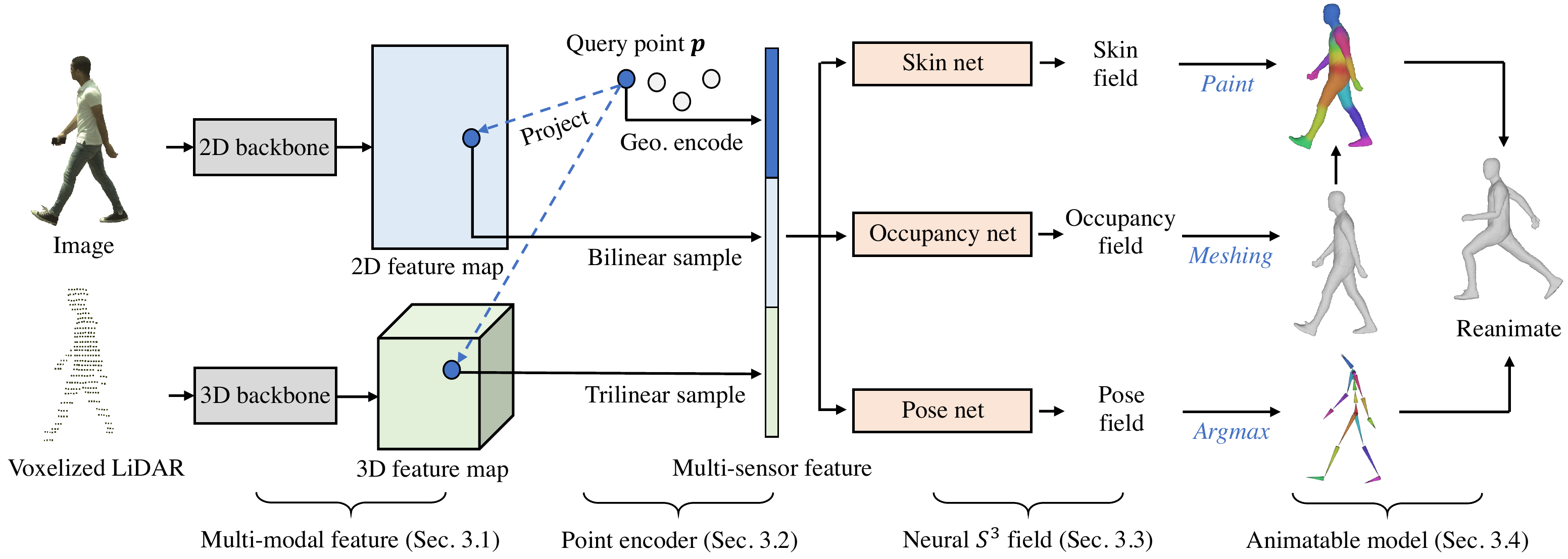}
\end{center}
\cutcaptionup
\caption{Overview of our proposed method. From left to right: We process input sensor data into spatial feature representations (Sec \ref{sec:feature}). We query points adaptively from 3D space and extract their point encoding (Sec. \ref{sec:encoder}), which we use to query our neural implicit representations of shape, pose, and skinning (Sec. \ref{sec:header}). We apply post-processing to construct the final explicit representation of an animatable person (Sec. \ref{sec:postprocess}).
}
\cutcaptiondown
\label{fig:overview}
\end{figure*}

\cutsectionup
\section{Reconstructing Animatable Pedestrians}
\cutsectiondown
We propose {\bf S$^3$}, an end-to-end neural {\bf S}hape, {\bf S}keleton, and {\bf S}kinning field model to reconstruct realistic and animatable 3D humans from  either a single camera image,  a single view LiDAR sweep, or a paired image and LiDAR input.
Assuming backgrounds are masked out for both image and LiDAR, our network takes image $\mathcal{I}$ and/or LiDAR $\mathcal{C}$ as input to form  $\mathbf{x} = (\mathcal{I}, \mathcal{C}) \in \mathcal{X}$. For a 3D query point in continuous space $\mathbf{p} \in \mathbb{R}^3$, the network outputs a continuous-valued multi-dimensional vector representing occupancy, joints positions as well as the skinning weights.

Fig.~\ref{fig:overview} depicts an overview of our approach.
We first encode the input $\mathbf{x}$  into 2D and 3D feature representations via a backbone feature network (Sec.~\ref{sec:feature}).
An implicit feature vector is then obtained for a query 3D point based on the generated feature representations (Sec.~\ref{sec:encoder}).
Given the feature vector  at a 3D location, its occupancy probability, human joint probability and skinning weight are predicted using a fully-connected network (Sec.~\ref{sec:header}).
Finally, shapes, skeletons and skinning weights are extracted based on the final predictions of the sampled points (Sec.~\ref{sec:postprocess}).

\cutsubsectionup
\subsection{Multi-Modal Feature}
\cutsubsectiondown
\label{sec:feature}
To extract multi-modal feature representations for animatable pedestrian generation, we process a sparse LiDAR point cloud and a single camera image through two separate backbone networks to obtain the 3D volumentric features and the 2D image features, respectively.

\cutparagraphup
\paragraph{Volumetric feature representation:}
The volumetric feature backbone takes as input the LiDAR data and outputs a dense 3D voxel tensor, which encodes the 3D shape information given from the sparse point cloud.
Specifically, we first transform the input point cloud $\mathcal{C}$ into a canonical coordinate frame.
In particular, the  point cloud is rotated along the yaw axis and translated along horizontal plane,  such that its yaw angle and horizontal translation is normalized to be zero-centered.
We then voxelize the normalized point cloud to a voxel grid.
A 3D convolution network $g_\mathrm{vox}$ is then exploited to produce a volumetric feature tensor $g_\mathrm{vox}(\mathcal{C})$.
Our 3D convolution network has a U-Net architecture~\cite{ronneberger2015u} due to its expressiveness and efficiency.
The encoder contains 8 convolution layers and the decoder contains 6 convolution layers with skip connections.
The final output feature has the same resolution as the input voxel grid.
Please see supplementary for more details.

\cutparagraphup
\paragraph{Image feature representation:}
To take advantage of the rich semantic and shape cues from images, we use a 2D convolution network to compute an image feature map $g_\mathrm{im}(\mathcal{I})$ for image $\mathcal{I}$.
The input image is ROI-cropped and masked by a given instance mask so that the target human is centered at the image with normalized size and the background is clean.
Our 2D convolution network is a four-stacked hourglass network~\cite{newell2016stacked} with the modifications proposed in~\cite{saito2019pifu}.
The final output feature has $1/4$ spatial resolution compared  to the input image.

\cutsubsectionup
\subsection{Point Feature Encoding}
\cutsubsectiondown
\label{sec:encoder}

To better predict a continuous multi-dimensional neural field representation of shape, pose, and skinning we  exploit  contextual information encoded from the multi-modal input and the inductive biases from its 3D positions w.r.t. the observer.
This point feature is an aggregation of the three features, namely the voxel, image and viewpoint features.
\begin{align}
	\phi(\mathbf{p}, \mathbf{x}) = [
	\phi_\mathrm{vox}(\mathbf{p}, \mathcal{C}),
	\phi_\mathrm{im}(\mathbf{p}, \mathcal{I}),
	\phi_\mathrm{view}(\mathbf{p})]
	\label{eqn:implicit_feat}
\end{align}
We now describe these features in more detail.

\cutparagraphup
\paragraph{LiDAR and Image Encoding:}
Given the input voxel feature $g_\mathrm{vox}(\mathcal{C})$ and image feature $g_\mathrm{im}(\mathcal{I})$,
we compute the feature vector at an arbitrary continuous point $\mathbf{p}$ through interpolation.
Specifically, we use trilinear interpolation to obtain corresponding 3D feature, denoted as $\phi_\mathrm{vox}(\mathbf{p}, \mathcal{C}) = \texttt{TrilinearSample}(\mathbf{p}, g_\mathrm{vox}(\mathcal{C}))$;
We also project the point onto the camera plane and retrieve 2D image feature through bilinear interpolation, denoted as $\phi_\mathrm{im}(\mathbf{p}, \mathcal{I}) = \texttt{BilinearSample}(\pi(\mathbf{p}), g_\mathrm{im}(\mathcal{I}))$, where $\pi$ is the perspective projection function. Here we assume the calibration is known a priori.

\cutparagraphup
\paragraph{Viewpoint Encoding:}
To provide the global context, we  compute an additional global positional feature $\phi_\mathrm{view}(\mathbf p)$ for a given query point $\mathbf{p}$, expressing how $\mathbf p$ is viewed wrt the sensor viewing angle.
The viewpoint feature $\phi_\mathrm{view}(\mathbf p)$ is defined as the inner product between sample point $\mathbf p$ and the unit camera ray $\frac{\mathbf r}{\left\Vert \mathbf r \right\Vert_2}$ passing the origin $\mathbf c$:
\begin{align}
	\phi_\mathrm{view}(\mathbf p) = (\mathbf p - \mathbf c) \cdot \frac{\mathbf r}{\left\Vert \mathbf r \right\Vert_2}
	\label{eqn:depth_feat}
\end{align}
where the origin $\mathbf c$ is defined as the pelvis joint or centre of LiDAR points cloud.
This feature is especially useful in areas where the LiDAR is missing, the query points on the same outgoing camera ray will have same pixel feature $\phi_\mathrm{im}(\mathbf{p}, \mathcal{I})$, this viewpoint encoding can help to distinguish those sample points.

\cutsubsectionup
\subsection{Neural S$^3$ Field}
\cutsubsectiondown
\label{sec:header}
In this section we describe how we predict the final output $f(\phi(\mathbf{p}, \mathbf{x}))$ for each query point $\mathbf p$.
The final output forms a high-dimensional vector field in continuous 3D space, encoding the occupancy, joint position and skinning weight for an arbitrary given point.
The input to the network $f$ is the point encoding $\phi(\mathbf{p}, \mathbf{x})$  described in Eqn.~\ref{eqn:implicit_feat}.
Our network consists of three $5$-layer multi-layer perceptrons (MLPs), namely the occupancy net $f_\mathrm{occ}$, the pose net $f_\mathrm{pose}$, and the skinning net $f_\mathrm{skin}$.
We use three separate  headers to improve the prediction capacity for each task without sharing weights, whereas their shared input feature still promotes consistency across each task's output.
We do not share the weights among the modules to allow for better expressiveness.

\cutparagraphup
\paragraph{Occupancy Net:} Inspired by  \cite{saito2019pifu,OccupancyNet}, our occupancy head outputs a probability value $\mathbf o \in [0, 1]$ representing the probability of the point $\mathbf p$ being inside the human.
\[f_\mathrm{occ}(\phi(\mathbf{p}, \mathbf{x})) \rightarrow \mathbf{o} \in [0, 1]\]
Such continuous representation is easier to train than implicit shape representations, such as the signed distance function, while maintaining expressive power. 

\cutparagraphup
\paragraph{Pose Net:}
The human skeleton is represented as a set of key-points and interconnected bones in 3D. To recover the underlying skeleton, the pose net  maps each query point $\mathbf p$ into a $M$-dimensional probability vector  $\mathbf j$ representing  the likelihood that the query point belongs to each human joint
\[f_\mathrm{pose}(\phi(\mathbf{p}, \mathbf{x})) \rightarrow \mathbf{j} \in [0, 1]^M\]
where $M$ is the number of joints.

\cutparagraphup
\paragraph{Skinning Net:}
Skinning is the process that binds the vertices on the 3D mesh to each bone in the human skeleton. For instance, rotating the neck of a character should most likely influence the vertices on the head.
The skinning weights describe the influence of each bone's rigid motion on a given vertex, and the values are normalized such that the sum of skinning weights on all bones equals to 1.
To reconstruct the skinning weights, we learn a skinning function over the human surface, which maps each surface point on the human mesh into a skinning vector.
Similar to the other two heads, our skinning net takes the point encoding $\phi(\mathbf{p}, \mathbf{x})$ as input and outputs a $K$-dimentional probability simplex,
\[f_\mathrm{skin}(\phi(\mathbf{p}, \mathbf{x})) \rightarrow \mathbf{s} \in \Delta^K\]
with $K$ the number of joints.
In addition, we add a softmax layer after the fully-connected layers of skinning head to make the prediction  sum to one.

\begin{figure}[t]
\begin{center}
	\includegraphics[width=0.45\textwidth]{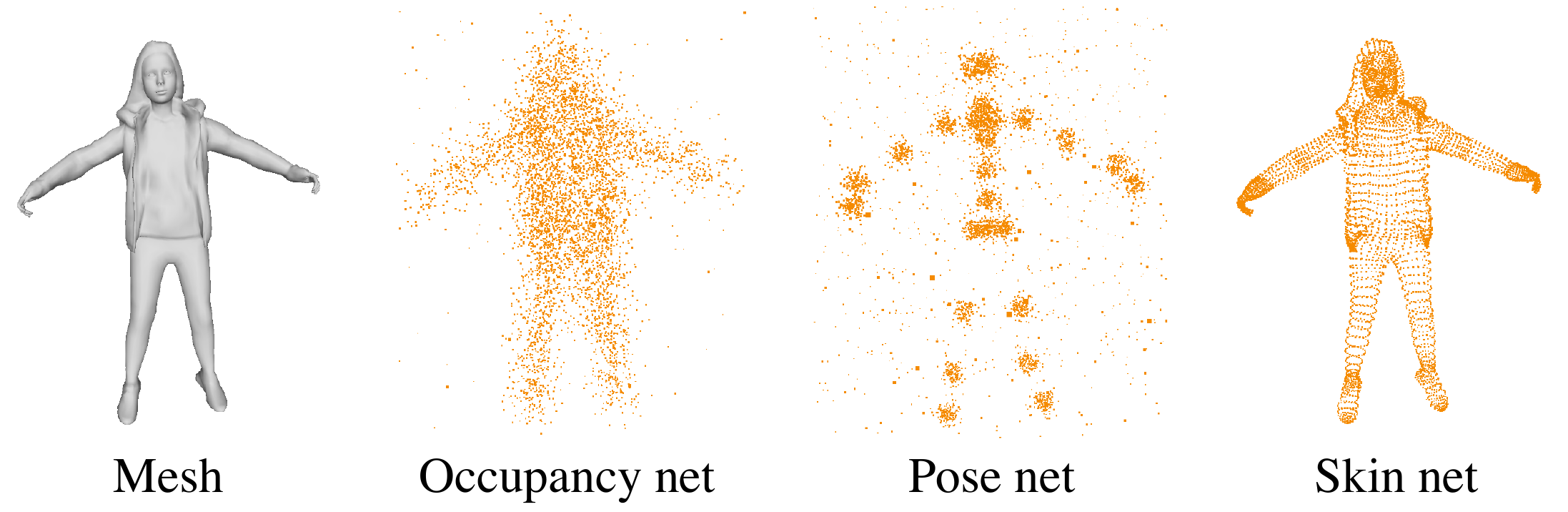}
\end{center}
\cuthalfcaptionup
\caption{Sampled points to compute loss on each module.}
\cuthalfcaptiondown
\label{fig:sample}
\end{figure}

\cutsubsectionup
\subsection{Animatable Model Extraction}
\cutsubsectiondown
\label{sec:postprocess}
Suppose now we have learned the function $f(\phi(\mathbf{p}, \mathbf{x}))$. The next step is to extract an animatable human model from it.
We first construct a dense grid of query points $\mathcal{P} = \{ \mathbf p_i \}_{i=1}^N$ according to the desired resolution.
For each query point $\mathbf p$, we evaluate the probaility of its volumetric occupancy $\mathbf o $ and body joint $\mathbf j$, creating a dense field of occupancy as well as {joint heatmaps}.
Then we run the Marching Cubes algorithm~\cite{lorensen1987marching} on the occupancy grid to generate the human mesh.
We also locate all the human joints' positions by taking its most probable location over each joint's heatmap.
Finally, for each vertex $\mathbf v$ of the reconstructed mesh, we compute the per-vertex skinning weight  $\mathbf s$.
Fig.~\ref{fig:overview} (right) depicts such animatable model extraction process.

\cutsubsectionup
\subsection{Network Training}
\cutsubsectiondown
\label{sec:learning}
We learn our implicit functions parameterized by neural network weights through a dataset composed of pairs of sensory input and ground-truth animatable mesh.
The networks parameters are trained by jointly minimizing the following loss function:
\begin{align}
	L = \lambda_{\mathrm{occ}} L_{\mathrm occ} + \lambda_{\mathrm pose} L_{\mathrm pose} + \lambda_{\mathrm skin} L_{\mathrm skin}
	\label{eqn:all_loss}
\end{align}
Fig.~\ref{fig:sample} shows the sampled query points to train the three implicit functions. The parameters of all the backbone and head networks in $f(\phi(\mathbf{p}, \mathbf{x}))$ are learned jointly through back-propagation over the sampled query points.
We now describe the loss functions and query point sample strategy for each head in detail.

\begin{figure*}[t]
\begin{center}
	\includegraphics[width=0.95\textwidth]{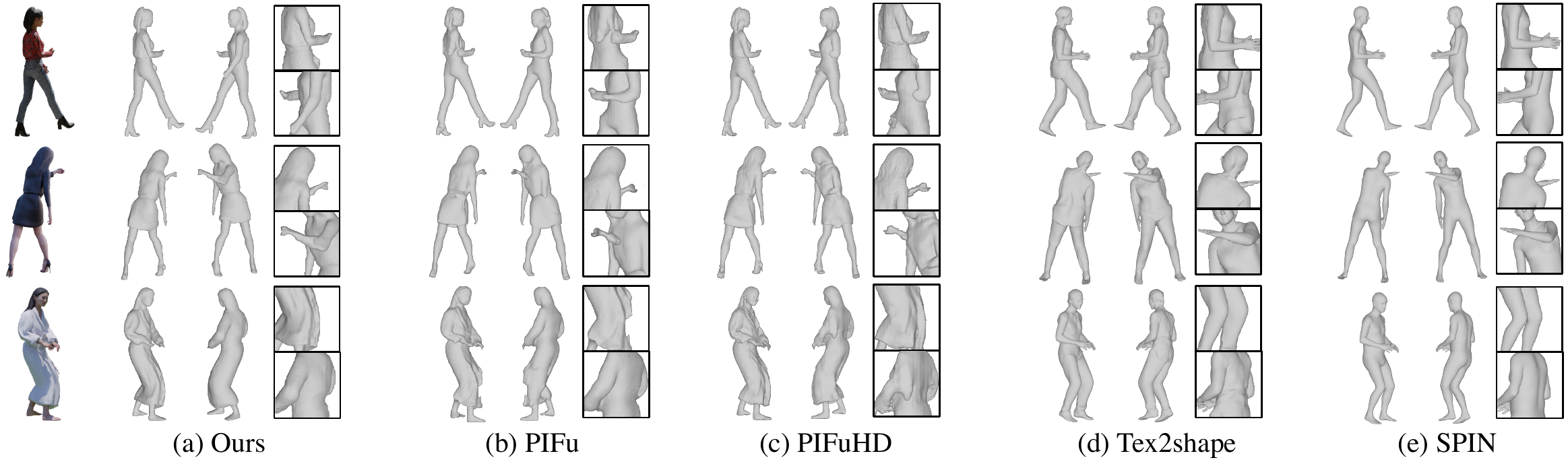}
\end{center}
\cutcaptionup
\caption{Reconstruction on RenderPeople dataset. In each cell, we show the shape in front, invisible, and zoomed-in views.}
\cutcaptiondown
\label{fig:surface_recon_sim}
\end{figure*}

\cutparagraphup
\paragraph{Occupancy Loss:}
This loss  encodes the disagreement between the ground truth (GT) occupancy  and the estimated occupancy probability.
During training, we sample a batch of query points $\mathbf p_i \in \mathbb R^3$, $i\in 1, 2, \cdots N_\mathrm{occ}$ from the continuous 3D space, and use
the mean square error between the ground-truth occupancy $o_i$ and
predicted occupancy $f_\mathrm{occ}(\phi(\mathbf{p}_i, \mathbf{x}))$ as occupancy loss function:
\begin{align}
	L_{\mathrm occ} = \frac{1}{N_\mathrm{occ}} \sum_{i=1}^{N_\mathrm{occ}} \left( \mathbf{o}_i - f_\mathrm{occ}(\phi(\mathbf{p}_i, \mathbf{x})) \right)^2
	\label{eqn:occupancy_loss}
\end{align}
We follow~\cite{saito2019pifu} and sample near surface points and uniform points in the space as the query {training} points.
For near surface points, we randomly sample points from the mesh surface, and perturb them with Gaussian noise.
Such sampling strategy ensures training efficiency while still learning a high-definition surface boundary.

\cutparagraphup
\paragraph{Pose Loss:}
We use the mean square error between the ground-truth key-point probability $\mathbf{j}_i$
and predicted joint probability $f_\mathrm{pose}(\phi(\mathbf{p}_i, \mathbf{x}))$ as the pose loss:
\begin{align}
	L_{\mathrm pose} = \frac{1}{N_\mathrm{pose}} \sum_{i=1}^{N_\mathrm{pose}} \left( \mathbf{j}_i - f_\mathrm{pose}(\phi(\mathbf{p}_i, \mathbf{x})) \right)^2
	\label{eqn:pose_loss}
\end{align}
where the GT key-point probability $\mathbf{j}_i$ is determined by computing the Gaussian function on the distance between query point $\mathbf p_i$ and GT key-point location.
Unlike prior 2D and 3D pose estimation methods which learn a joint probability on a dense grid heatmap~\cite{newell2016stacked,sun2019deep}, we sample the query points $\mathbf p_i$ from a Gaussian distributions centered at the GT key-point locations, and we augment them with uniformly sampled points in the space to ensure background points are also covered. Fig.~\ref{fig:sample} depicts an example of such sampling strategy.

\cutparagraphup
\paragraph{Skinning Loss:}
For each queried point $\mathbf p_i$ during training, we also apply the mean square error between the ground-truth skinning weight $\mathbf{s}_i$
and predicted skinning weight $f_\mathrm{skin}(\phi(\mathbf{p}_i, \mathbf{x}))$ as the loss function:
\begin{align}
	L_{\mathrm skin} = \frac{1}{N_\mathrm{skin}} \sum_{i=1}^{N_\mathrm{skin}} \left( \mathbf{s}_i - f_\mathrm{skin}(\phi(\mathbf{p}_i, \mathbf{x})) \right)^2
	\label{eqn:skinning_loss}
\end{align}
{where $N_\mathrm{skin}$ is the number of query points}.


\cutsectionup
\section{Experimental Evaluation}
\cutsectiondown
In this section, we begin by introducing our experimental setting. 
We then compare our approach against several state-of-the-art methods in 3D shape reconstruction and present some ablation studies.
Due to the page limits, we refer the readers to see the additional ablation studies in the supplementary material.
We also showcase our method's ability to animate the reconstructed human shape to target poses.

\cutsubsectionup
\subsection{Datasets}
\cutsubsectiondown
We train our algorithm on the RenderPeople~\cite{renderpeople} dataset, where ground-truth shapes are available. We evaluate our model on held out RenderPeople characters and animations as well as  on a large-scale self-driving dataset containing pedestrians with large variations in pose, shape, clothing, under diverse environmental illuminations.

\cutparagraphup
\paragraph{RenderPeople Dataset:}
We purchased $793$ RenderPeople rigged characters~\cite{renderpeople} and then animate them with $39$ different animations from Mixamo~\cite{mixamo}. 
Our character set contains a wide array of pedestrians with different body shapes and clothing.
We extend the diverse set of 32 animations from ~\cite{li2020monocular} with 7 additional animations that cover  pedestrian behaviors, such as walking and running.
We randomly select $3$ frames per animation for each model, resulting in $777 \times 39 \times 3 = 90,909$ meshes for training, and $16 \times 39 \times 3 = 1,872$ meshes for evaluation.
In addition, we use an extra 10 novel animations from Mixamo (result in $16 \times 10 \times 3 = 480$ meshes) to test the generalization ability of our model to unseen animations.
To create realistic images, we use Blender Cycles engine~\cite{blender} to render images.
Characters and placed at a depth of 10 meters and view angle is uniformly distributed around yaw angle. We use $\approx 400$ outdoor HDRI image from HDRI Haven~\cite{hdri} for environment lighting.  In addition, we use Intel Embree ray tracer~\cite{wald2014embree} to simulate LiDAR point cloud. Random ray dropping and disturbance are injected to push realism.
We generate training and testing examples every 20 degrees and 120 degrees around the yaw-axis.
Please see more details in supplementary.

\cutparagraphup
\paragraph{Self-driving Dataset:}
We also conduct experiments on a real-world self-driving dataset, which contains diverse scenes across multiple metropolitan cities in North America.
Our self-driving sensor platform has a global-shutter camera with a $112^\circ$ horizontal FOV and a 64-beam spinning LiDAR.
To create our dataset, we label 1.3k pedestrians that are within 6-25m of the ego-car with 3D box annotations and image-based instance annotations, resulting in over 1k evaluation examples.

\begin{table}[ht]
    \begin{center}
    \scalebox{0.95}{
    \begin{tabular}{lccccc}
	\toprule[0.1em]
	& \multicolumn{3}{c}{Seen Poses} & \multicolumn{2}{c}{Unseen Poses}  \\
    & CD $\downarrow$ & P2S$\downarrow$ & Norm$\uparrow$ & CD $\downarrow$ & P2S$\downarrow$  \\
    \midrule
    SPIN$^{\ast}$~\cite{kolotouros2019learning}  & 3.25 & 2.97 & 0.76 &  3.26&  2.97  \\
    PIFu$^{\ast, \dagger}$~\cite{saito2019pifu}  & 2.78 & 2.63 & 0.78 &  2.74 & 2.58\\
    PIFuHD$^{\ast, \dagger}$~\cite{saito2020pifuhd}  & 2.47 & 2.44 & 0.79 & 2.41 & 2.38\\
    PIFu~\cite{saito2019pifu}   & 1.03 & 1.05 & 0.88  &  1.61 & 1.70\\
    \midrule
    Ours (img)  & 0.92 & 0.93 & 0.89  & 1.58  & 1.63\\
    Ours (img+lidar) & \textbf{0.66} & \textbf{0.65} & \textbf{0.91}  &  \textbf{0.76} & \textbf{0.75}\\
	\bottomrule[0.1em]
    \end{tabular}
    }
    \end{center}
    \cuthalftablecaptionup
    \caption{Quantitative comparisons on RenderPeople dataset. $^\ast$: model pre-trained from other RenderPeople dataset; $^\dagger$: model using orthographic images. We report the Chamfer error and P2S error in cm for both seen and unseen poses respectively, and the lower is better. For Normal consistency, the higher is better.}
    \cuthalftablecaptiondown
    \label{tab:system_render_people}
\end{table}

\begin{figure*}[t]
\begin{center}
	\includegraphics[width=0.95\textwidth]{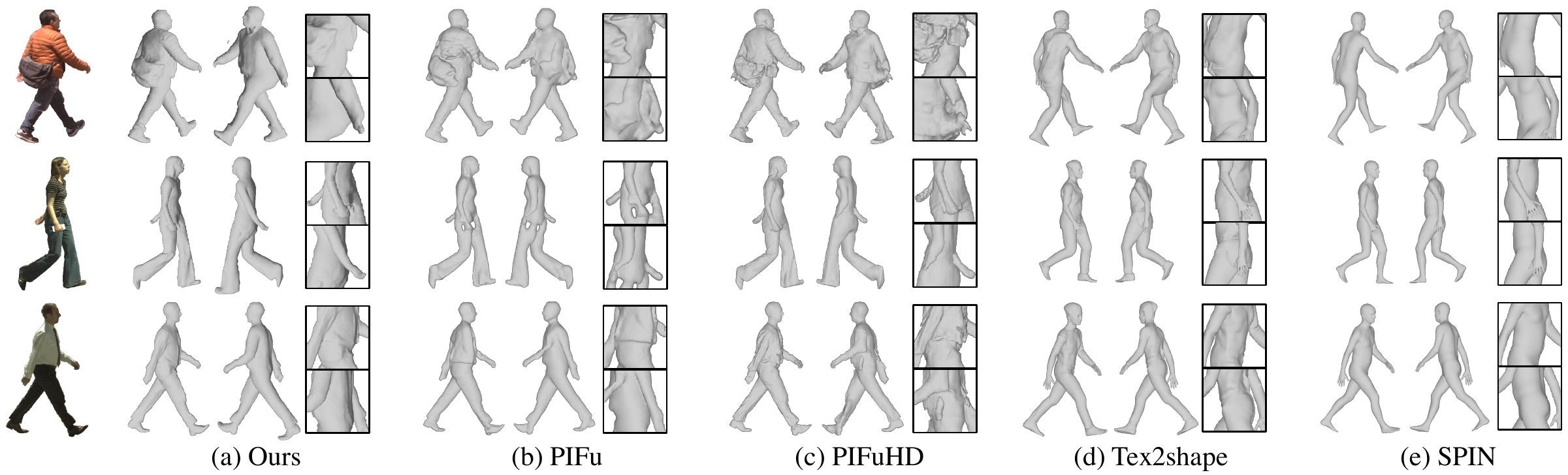}
\end{center}
\cutcaptionup
\caption{Reconstruction on real-world data. In each cell, we show the shape in front, invisible, and zoomed-in views.}
\cutcaptiondown
\label{fig:surface_recon}
\end{figure*}

\begin{figure}[t]
\begin{center}
	\includegraphics[width=0.49\textwidth]{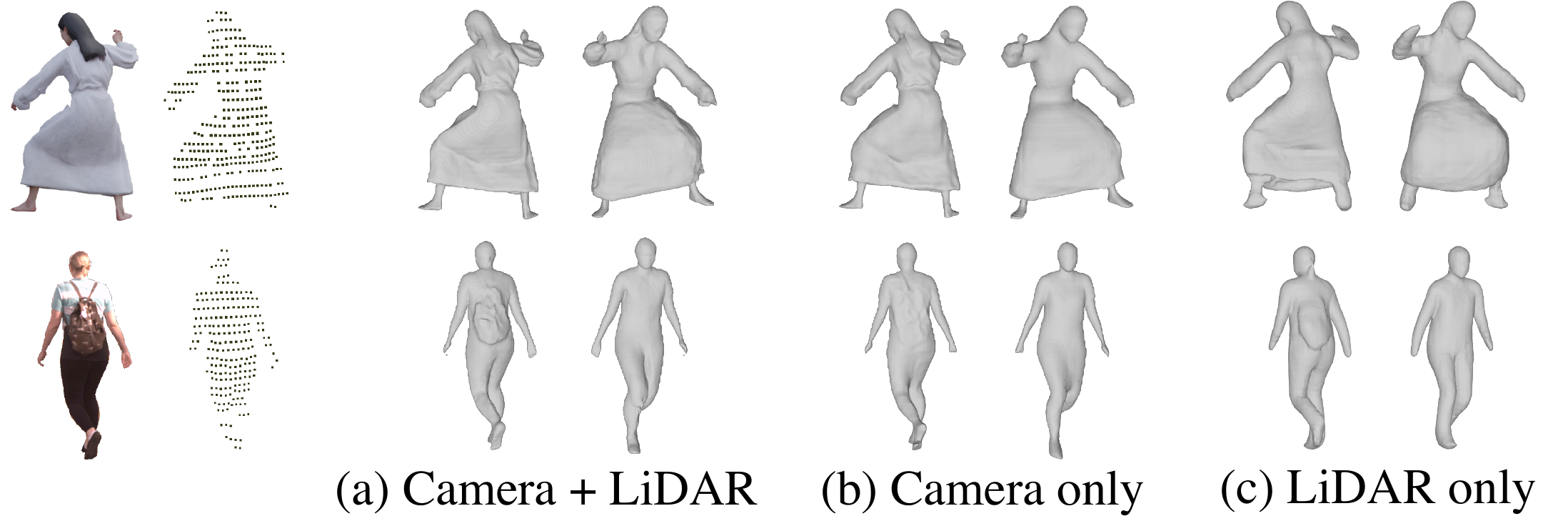}
\end{center}
\cutcaptionup
\caption{Reconstruction with different sensory inputs.
}
\cutcaptiondown
\label{fig:multi_sensor_surface}
\end{figure}

\cutsubsectionup
\subsection{Experimental Details}
\cutsubsectiondown

\paragraph{Implementation Details:}
We train our model on $512 \times 512$ RenderPeople images using 16 GPUs with a batch size of $64$ for $14$ epochs. 
We voxelize the LiDAR sweep to a voxel grid with shape $64 \times 64 \times 64$.
We use RMSProp optimizer with initial learning rate of $1\times 10^{-3}$, and then decay it by 10 at 10th epochs and 12th epochs.
We sample 5,000 points for each module to compute the loss function during training.
During inference, we evaluate the neural implicit fields to construct a $256^3$ spatial resolution volume and perform Marching Cubes with iso-surface threshold at $0.5$ to extract the output meshes. 

\cutparagraphup
\paragraph{Baselines:}
We compare our model with several state-of-the-art shape reconstruction approaches, namely PIFu~\cite{saito2019pifu}, PIFuHD~\cite{saito2020pifuhd}, Tex2Shape~\cite{alldieck2019tex2shape} and SPIN~\cite{kolotouros2019learning}.
Due to the lack of training scripts or configurations, we directly use the officially released models that are trained on the authors' private RenderPeople datasets, which might differ from ours in terms of poses, characters, lighting and rendering parameters. 
Note that the pretrained model provided by PIFu~\cite{saito2019pifu} and PIFuHD~\cite{saito2020pifuhd} are trained on orthographic camera images.
We found they do not transfer to perspective camera well. 
To ensure a fair comparison during testing, we render orthographic camera images as input for these models while keeping the remaining rendering settings constant, including character, action, camera pose, lighting, and rendering engine.
There is also a global coordinate/pose shift between the shape predicted by the pre-trained models and our GT. 
We thus perform point-to-plane ICP \cite{chen1992object} between the predicted and ground-truth shape before evaluation.
We also scale SPIN~\cite{kolotouros2019learning}'s predicted meshes such that they are under the same scale as our GT meshes.
In addition, we also train PIFU model from scratch on the same training set as ours. 

\cutparagraphup
\paragraph{Metrics:}
We evaluate the reconstruction performance of our model on three metrics: (1) an average Chamfer distance (cm) between our reconstructed mesh and the GT mesh, (2) an average point-to-surface (P2S) distance (cm) between the vertices on our reconstructed surface to the GT surface, and (3) a normal consistency measure, which is defined as the average of \textit{normal accuracy} and \textit{normal completeness}.
We compute the normal accuracy by sampling 10,000 points from the predicted mesh surface, finding their closest points on the ground-truth mesh surface, and computing the cosine similarity between their normal directions.
We compute the normal completeness by sampling 10,000 points from the GT mesh surface, finding closest points on the prediction surface, and computing the cosine similarity between their normal directions.

\begin{table}[t]
    \begin{center}
    \begin{tabular}{cccc}
	\toprule[0.1em]
    Sensor & Chamfer$\downarrow$ & P2S$\downarrow$ & Normal$\uparrow$ \\
    \midrule
    Image & 0.92 & 0.93 & 0.89 \\
    LiDAR & 1.16 & 1.30 & 0.89 \\
    Image+LiDAR & \textbf{0.66} & \textbf{0.65} & \textbf{0.91} \\
	\bottomrule[0.1em]
    \end{tabular}
    \end{center}
    \cuthalftablecaptionup
    \caption{Ablation on multi-sensor feature.}
    \cuthalftablecaptiondown
    \label{tab:multi_sensor}
\end{table}

\begin{figure}[t]
\begin{center}
	\includegraphics[width=0.35\textwidth]{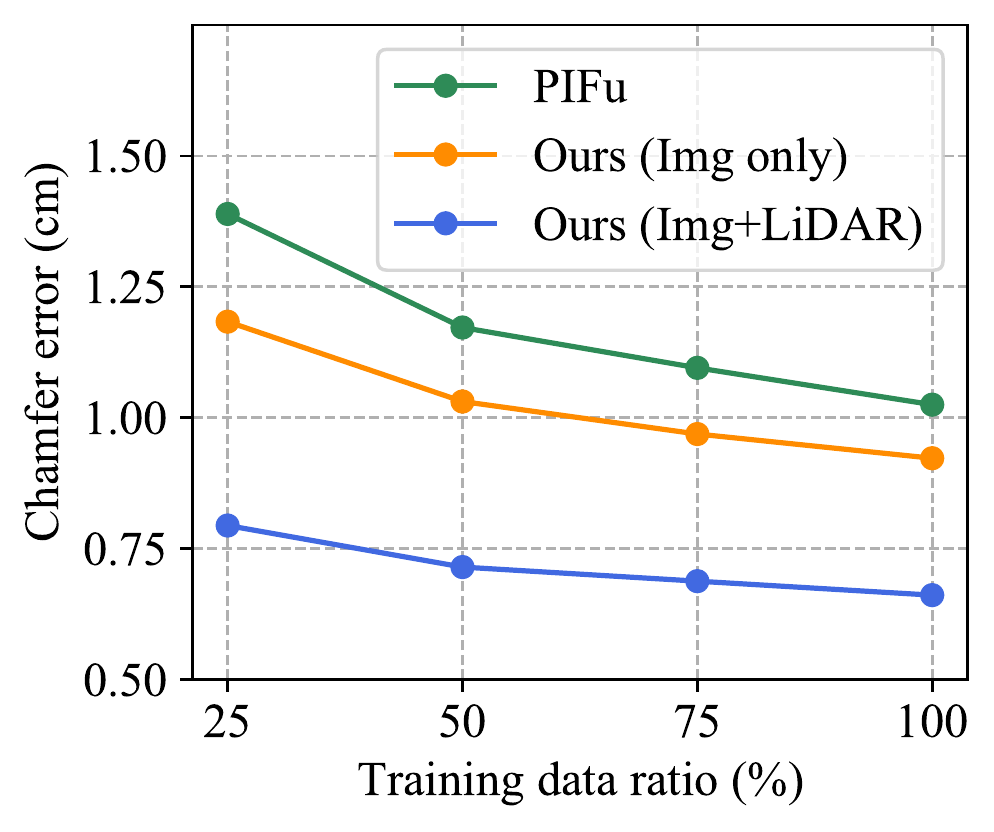}
\end{center}
\cuthalfcaptionup
\caption{Chamfer error as a function of training set scale.} 
\cuthalfcaptiondown
\label{fig:ablate_dataset_scale}
\end{figure}

\cutsubsectionup
\subsection{RenderPeople Dataset Results}
\cutsectiondown

\paragraph{State-of-the-art (SoTA) Comparison:}
We compare our model with SOTA on our test set (RenderPeople). 
We use $^\ast$ to denote pre-trained models on third-party datasets and $^\dagger$ to represent the usage of orthographic camera images.
As shown in Table~\ref{tab:system_render_people},  our method achieves the best performance in all metrics.
Note that PIFu and PIFu$^{\ast, \dagger}$ have a significant performance gap, as the RenderPeople datasets and data generation process may differ in terms of pose distribution, lighting, rendering engine and  training dataset scale.
The PIFu model is equivalent to our image-only baseline using depth feature as viewpoint encoding.
Figure~\ref{fig:surface_recon_sim} shows a qualitative comparison. We show the result using our model (image+LiDAR) by default in all figures. Our model captures better topological structures and fine  details, especially on occluded parts.

\cutparagraphup
\paragraph{Quantitative results on unseen animations:}
Prior works~\cite{li2020monocular,saito2019pifu, alldieck2019learning, alldieck2019tex2shape} utilize a similar human pose distribution during training and testing. 
To test model generalization, we also evaluate reconstruction performance on models with 10 unseen animations.
As shown in Table~\ref{tab:system_render_people}, both our image only method and multi-modal method outperform all the competing approaches in this setting as well. 
It is worth noting that testing on novel pose/actions in general will hurt the performance for the methods trained on our training set, especially for image-only approaches.
This is because single-view 3D reconstruction is an ill-posed problem and there may be multiple solutions. 
Learning-based methods tend to pick solutions that are within the training set distribution.

\begin{figure*}[t]
\begin{center}
	\includegraphics[width=0.98\textwidth]{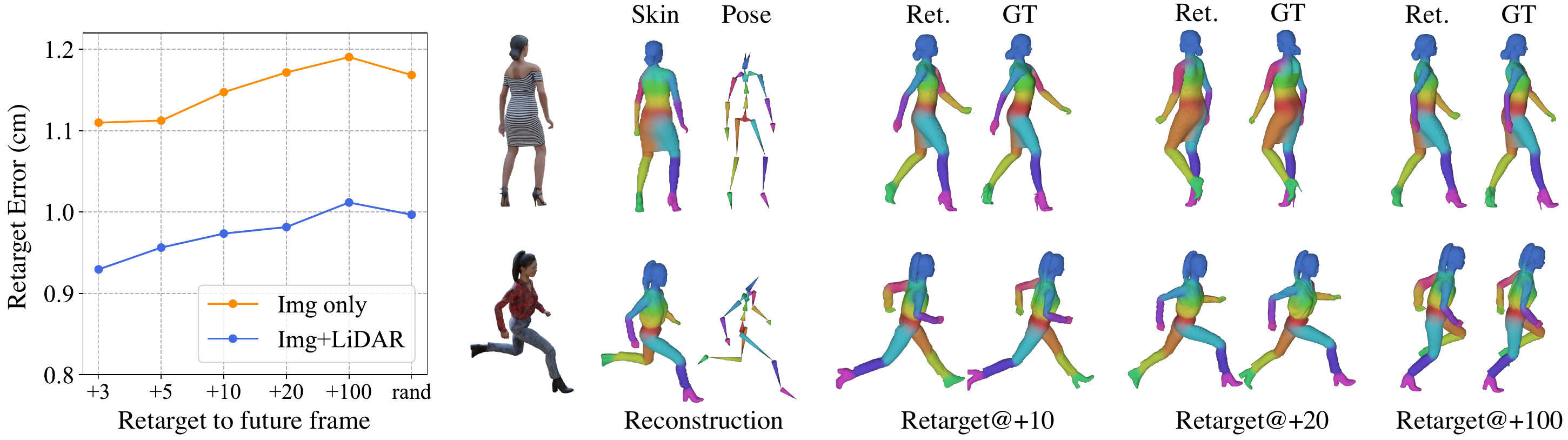}
\end{center}
\cutcaptionup
\caption{Reanimation results. Left: quantitative evaluation; Right: qualitative evaluation, from left to right: our predicted mesh, skeleton, mesh retarget to future ($10, 20, 100$) frames and GT mesh. In each cell, left is retarget shape, right is GT shape.}
\cutcaptiondown
\label{fig:animation_recon_sim}
\end{figure*}

\begin{figure*}[t]
\begin{center}
	\includegraphics[width=0.9\textwidth]{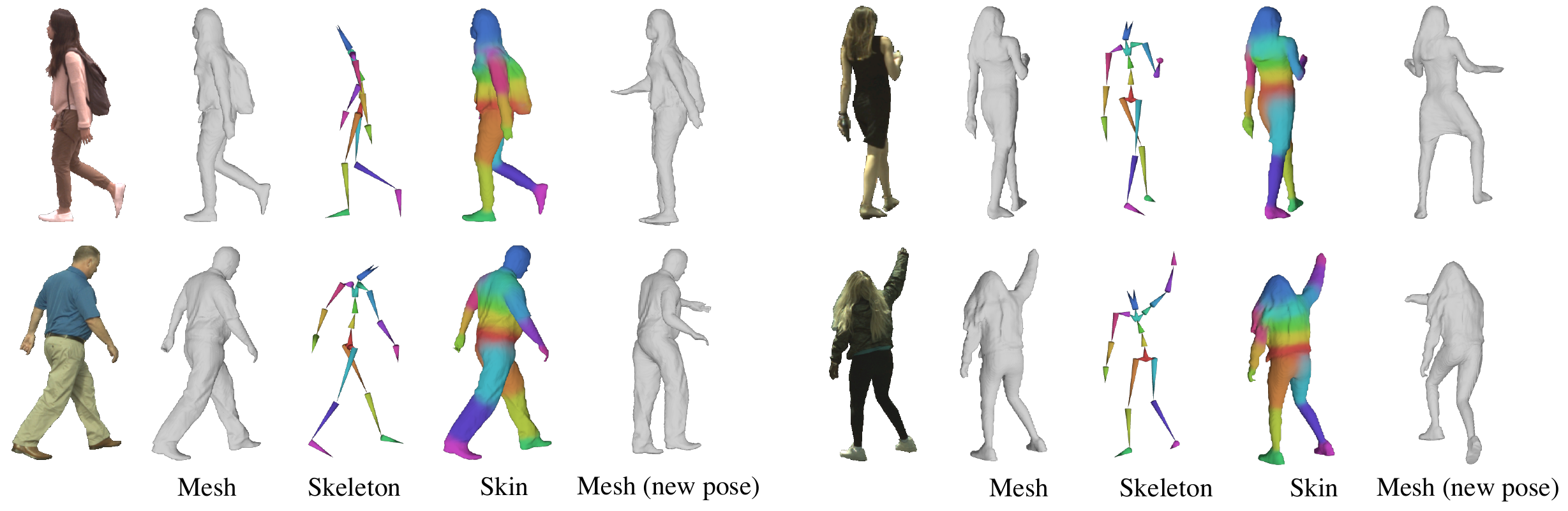}
\end{center}
\cutcaptionup
\caption{Our reconstructed mesh, skeleton and skinning weights from self driving data, which can be used to animate the mesh to novel poses.}
\cutcaptiondown
\label{fig:animation_recon_real}
\end{figure*}

\cutparagraphup
\paragraph{Ablation on sensor type:}
As shown in Table~\ref{tab:multi_sensor}, using all sensors achieves the best performance.
Figure~\ref{fig:multi_sensor_surface} shows  qualitative comparison.
While using sparse LiDAR alone captures global structure and human topology, the image provides additional information to refine the pose and shape details such as hair and clothing.

\cutparagraphup
\paragraph{Ablation on dataset scale:}
We also ablate the human reconstruction performance as a function of the training set size. As shown in Figure~\ref{fig:ablate_dataset_scale}, our approach performs well even when trained with only 25\% of the data. 

\cutsubsectionup
\subsection{Self-driving  Results}
\cutsubsectiondown
We qualitatively evaluate our method on self-driving data to reconstruct in-the-wild pedestrians in Figure~\ref{fig:surface_recon}.
Our reconstructions have less artifacts and better capture global topology, while PIFu~\cite{saito2019pifu} and PIFuHD~\cite{saito2020pifuhd} tend to either miss regions or predict incorrect body parts.
Our model is robust to light accessories although our training data does not include those accessories.

\cutsubsectionup
\subsection{Animating the Reconstructed  Characters}
\cutsubsectiondown
We now showcase how the reconstructed animatable 3D human model can be retargeted to  new poses.
Given an inferred 3D human mesh, skeleton and skinning weights, we use inverse kinematics (IK) to compute the transformation between our reconstructed skeleton and the target skeleton.
We then re-animate the mesh using linear blend skinning (LBS) through the inferred skinning weights.
Please see the supplementary for more details about our IK solver and the LBS model.

\cutparagraphup
\paragraph{Quantitative evaluation:}
We introduce a new metric to measure the overall performance of our model, denoted as retarget error.
Specifically, given a target human pose, we retarget our predicted human to generate the shape at this target pose, and we evaluate the chamfer distance (cm) between our re-animated shape and GT shape at the target human pose.
This retarget error measures the overall performance of surface reconstruction, pose estimation and re-animation of our model.
We conduct the evaluation on our RenderPeople test set, and we report the retarget error on both future frame ($+3, +5, +10, +20, +100$) and random frames in the animation.
As shown in Figure~\ref{fig:animation_recon_sim} (left), the error increases when retargeting to further frames, and it saturates at about $100$ future frames.
To the best of our knowledge, we are the first to quantitative study the human re-animation performance. We hope this will result in further work in the field.

\cutparagraphup
\paragraph{Qualitative evaluation:}
Figure~\ref{fig:animation_recon_sim} (right) shows a comparison of our retargeted mesh and GT mesh on RenderPeople dataset. 
We color the mesh with the skinning weights. 
We observe that the retargeting and skinning prediction are accurate, demonstrating the effectiveness of our model using a simple feed-forward network.
Figure~\ref{fig:animation_recon_real} shows the predicted mesh, skeleton, skinning weights visualizations, and the re-animated mesh in the target pose for examples in our self-driving dataset.
The re-posed meshes look realistic and preserve local details.

\begin{figure}[t]
\begin{center}
	\includegraphics[width=0.5\textwidth]{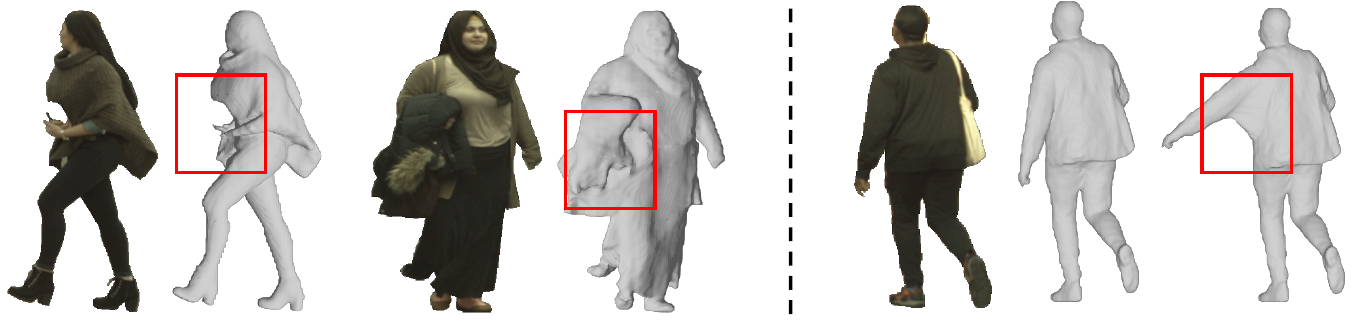}
\end{center}
\cuthalfcaptionup
\caption{Visualizations of failure cases. Our shape model failed due to inaccurate segmentation, severe accessories. Our animation model failed due to self-contacts, \etc.}
\label{fig:failure_case}
\cuthalfcaptiondown
\end{figure}

\cutparagraphup
\paragraph{Failure case analysis}
We also show the failure cases in Figure~\ref{fig:failure_case}. Our shape reconstruction module will fail due to segmentation error; severe clothing and accessories, \etc. 
This might because our training data does not include accessories. Our animation module will introduce artifacts in regions with self-contacts due to heavy clothing/accessories.

\cutsectionup
\section{Conclusion}
\cutsectiondown
In this paper, we propose S$^3$, a novel deep animatble human shape reconstruction algorithm.
Our network can take either a single image, a single LiDAR sweep, or both as input and predict a continuous multi-dimensional vector field in 3D space, representing occupancy, skeleton, and skinning weight. 
Our experimental results showcase that the proposed method achieves state-of-the-art performance in two challenging datasets. In addition, we demonstrate that we could generate 3D animated human character sequences using reconstructed 3D shapes from the proposed method.

{\small
\bibliographystyle{ieee_fullname}
\bibliography{egbib}
}

\clearpage
\section*{Appendix}
\appendix
\renewcommand{\thesection}{A\arabic{section}}  

In this supplementary material, we provide additional details, analyses and qualitative results of our method. We first describe the details about our network architecture and dataset preprocessing in Section~\ref{sec:additional_details}.
Then in Section~\ref{sec:additional_ablation} we show the additional ablation study on the RenderPeople dataset.
Next, we present more details about our animation formulation in Section~\ref{sec:animation_details}, including the Inverse Kinematics (IK) solution that transfers the reconstructed skeleton to the target skeleton, and the Linear Blend Skinning (LBS) formulation used for mesh re-animation.
Finally, we showcase more qualitative results in Section~\ref{sec:additional_results}.

\cutsectionup
\section{Additional Details}
\cutsectiondown
\label{sec:additional_details}

\begin{figure*}[ht]
\begin{center}
	\includegraphics[width=1.0\textwidth]{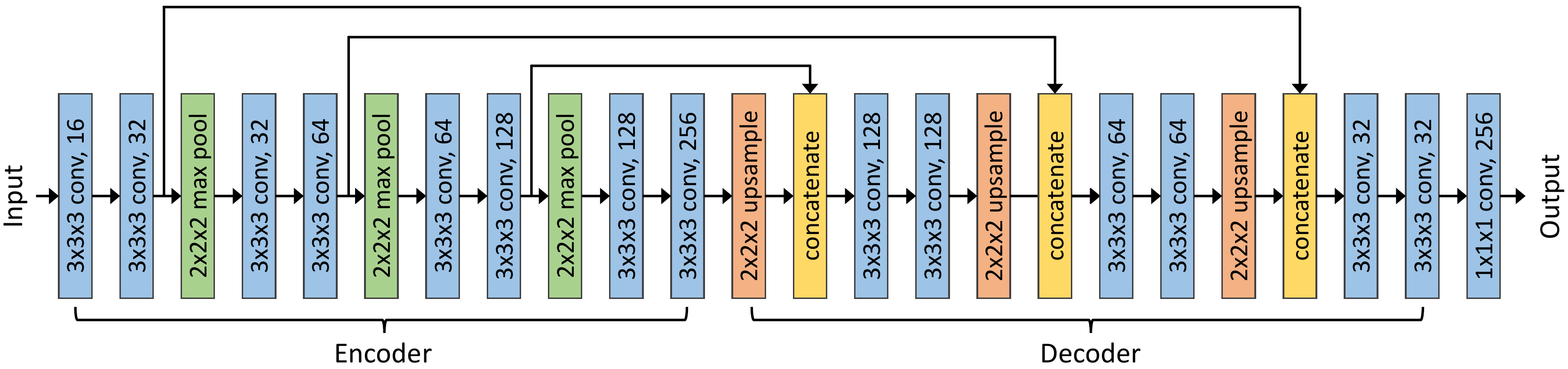}
\end{center}
\cuthalfcaptionup
\caption{Network architecture of the 3D U-Net.} 
\cuthalfcaptiondown
\label{fig:unet}
\end{figure*}

\paragraph{3D U-Net Architecture:}
We adopt the 3D U-Net architecture from~\cite{cciccek20163d}, where the encoder consists of 8 convolution layers. We gather intermediate features and place a max-pooling layer after every two conv layers with $stride=2$.
The decoder consists of 6 convolution layers. We fuse the intermediate features and place an upsampling layer after every two conv layers with $stride=2$.
We pass the resulting feature through one conv layer and obtain the final output. It has the same spatial resolution as the input. Please see Figure~\ref{fig:unet} for more details.

\cutparagraphup
\paragraph{Dataset Pre-processing:}
We purchase $793$ RenderPeople rigged characters~\cite{renderpeople} and animate them with $39$ different animations downloaded from Mixamo~\cite{mixamo}. See Table~\ref{tab:seen_list} for a complete list of our selected animation. 
We randomly select $3$ frames per animation for each model, where the sampled frames may be different across characters.
To test the generalization ability of our model, we use another $10$ \emph{held out} animations from Mixamo~\cite{mixamo}. Please see Table~\ref{tab:unseen_list} for more details.
We randomly select one HDRI image and rotate the light direction during the rendering.

There are three factors that will affect the size of a character in the rendered image: (1) the depth of the character, (2) camera intrinsics, and (3) the size of the character.
We fix the character depth at 10 meters and randomly perturb the camera's position tangent to its viewing direction to push characters away from the optical axis center. 
This simulates the severe perspective distortion observed in real images, reducing the sim-to-real gap. 
We also adjust the focal length to make sure the height of the rendered character is roughly equivalent to $90\%$ of the image. 
The output image resolution is $512 \times 512$ pixels.
To generate the corresponding LiDAR points of a 64-beam LiDAR sensor, we ray-cast the posed mesh at a 0.18$^{\circ}$ azimuth interval and about 0.4$^{\circ}$ elevation interval from $[2^{\circ}, -24^{\circ}]$ using Intel Embree~\cite{wald2014embree} ray tracing kernel.
To simulate ray-drop effect and sensor noise, we randomly drop $10\%$ ray-casted points and perturb each ray-casted points with a Gaussian noise along the ray direction ($\sigma=1$cm).

\begin{table}[ht]
    \begin{center}
    \begin{tabular}{c|c}
	\toprule[0.1em]
	Agreeing & Bored \\
	Breakdance Ready &  Defeat\\
	Defeated & Dwarf Idle \\
	Female Tough Walk & Hands Forward Gesture \\
	Holding Idle &  Jogging \\
	Look Over Shoulder & Old Man Idle \\
	Patting & Pointing \\
	Put Back Rifle Behind Shoulder & Run Look Back \\
	Running Right Turn & Running \\
	Searching Files High & Shoulder Rubbing \\
	Standing Clap & Standing Greeting \\
	Standing Torch Idle 02 & Standing Turn 90 Right \\
	Standing Turn Left 90 & Stop Jumping Jacks \\
	Strut Walking & Talking On Phone \\
	Talking Phone Pacing & Talking \\
	Texting And Walking & Walking Backwards \\
	Walking Left Turn & Walking Turn 180 \\
	Walking While Texting & Walking \\
	Walking-2 & Yawn\\ 
	Yelling & \\
	\bottomrule[0.1em]
    \end{tabular}
    \end{center}
    \cuthalftablecaptionup
    \caption{List of animations to generate training/test set.}
    \cuthalftablecaptiondown
    \label{tab:seen_list}
\end{table}

\begin{table}[ht]
    \begin{center}
    \begin{tabular}{c|c}
	\toprule[0.1em]
	Drunk Idle Variation & Drunk Walk \\
	Jog In Circle &  Pacing And Talking On A Phone\\
	Picking Up Object & Right Turn \\
	Slow Run & Taunt \\
	Thankful & Wheelbarrow Walk \\
	\bottomrule[0.1em]
    \end{tabular}
    \end{center}
    \cuthalftablecaptionup
    \caption{List of animations to generate unseen test set.}
    \cuthalftablecaptiondown
    \label{tab:unseen_list}
\end{table}

\cutsectionup
\section{Additional Ablations}
\cutsectiondown
\label{sec:additional_ablation}

\cutparagraphup
\paragraph{Ablation on point location encoding:}
We compare three different point location features.
(1) Compute the depth of the query points in camera coordinate and then normalize them to be zero centered. This can be viewed as the perspective version of PIFu~\cite{saito2019pifu}.
(2) Compute the point location of the query points in camera coordinate and then normalize them to be zero centered.
(3) Our proposed viewpoint encoding (Equation~2 in the main paper).
We note that in this ablation we use the image only model where the point location feature is more important.
As shown in Table~\ref{tab:geometric_encoding}, our proposed viewpoint encoding achieves the best performance. 
This suggests the importance of the viewpoint encoding $\phi_\mathrm{view}(\mathbf p)$ to disambiguate the query points lying on the same camera ray.

\begin{table}[t]
    \begin{center}
    \begin{tabular}{cccc}
	\toprule[0.1em]
    Geometry encoding & Chamfer$\downarrow$ & P2S$\downarrow$ & Normal$\uparrow$ \\
    \midrule
    depth & 1.025 & 1.045 & 0.882 \\
    point location & 0.935 & 0.952 & 0.886 \\
    ours & \textbf{0.922} & \textbf{0.928} & \textbf{0.891} \\
	\bottomrule[0.1em]
    \end{tabular}
    \end{center}
    \vspace{-10pt}
    \caption{Ablation on different point location encoding. Our viewpoint encoding performs best.}
    \label{tab:geometric_encoding}
\end{table}

\begin{figure}[ht]
\begin{center}
	\includegraphics[width=0.5\textwidth]{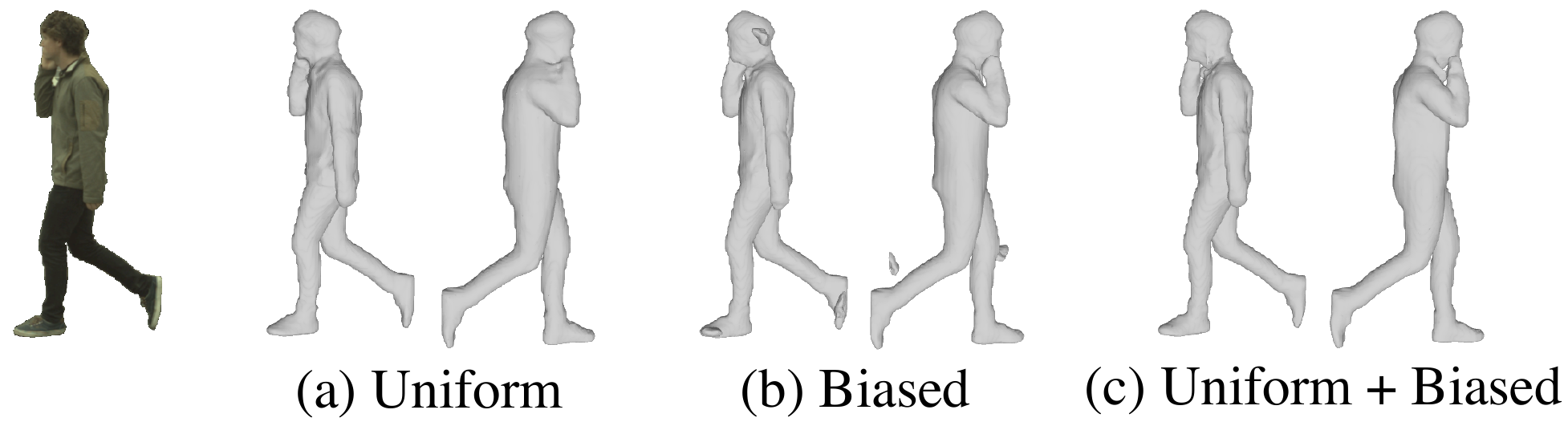}
\end{center}
\cuthalfcaptionup
\caption{Qualitative comparison of models trained with different sampling strategy. Using uniform sampling alone tends to have coarse and inaccurate surface. Using biased sampling alone tends to introduce artifacts outside the mesh.} 
\cuthalfcaptiondown
\label{fig:sample_strategy}
\end{figure}

\begin{figure*}[ht]
\begin{center}
	\includegraphics[width=0.92\textwidth]{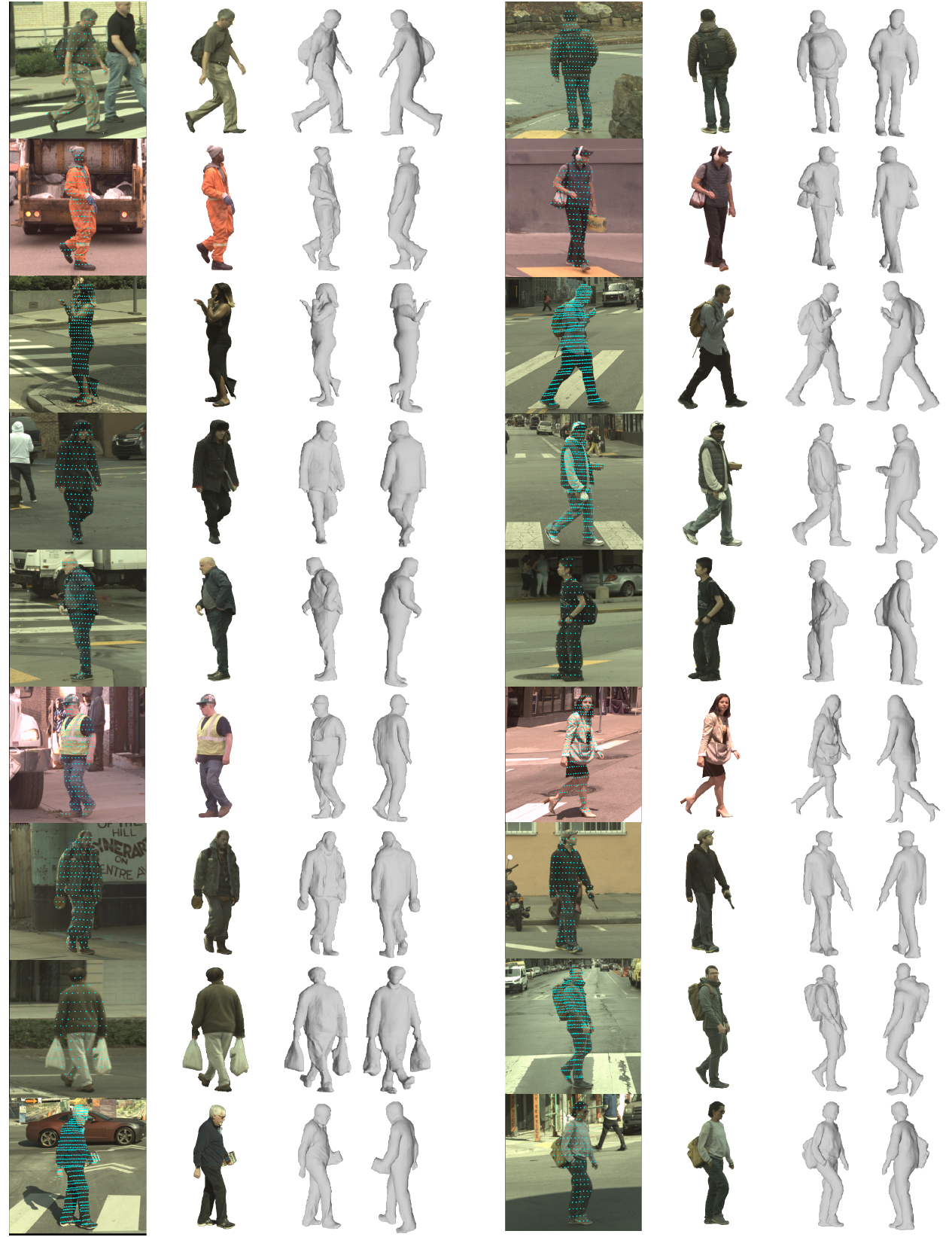}
\end{center}
\cuthalfcaptionup
\caption{Visualization of reconstruction on more urban scenes. From left to right in each cell: camera image overlaid with LiDAR points, foreground image, shape reconstruction.} 
\cuthalfcaptiondown
\label{fig:addition_recon}
\end{figure*}

\begin{figure*}[ht]
\begin{center}
	\includegraphics[width=0.92\textwidth]{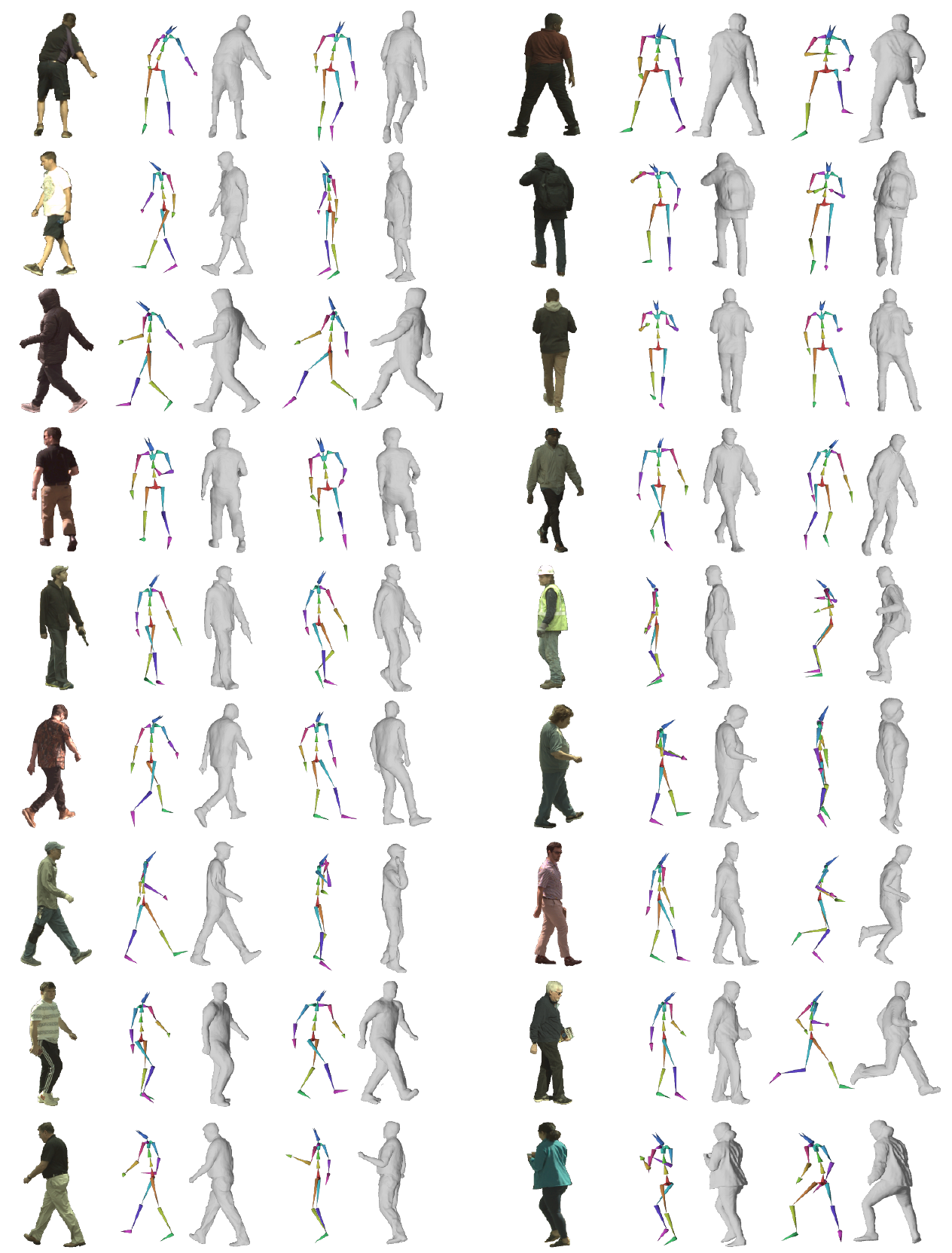}
\end{center}
\cuthalfcaptionup
\caption{Visualization of animation on more urban scenes. From left to right in each cell: foreground image, reconstructed skeleton, reconstructed shape, target skeleton, re-animated shape.} 
\cuthalfcaptiondown
\label{fig:addition_anim}
\end{figure*}

\cutparagraphup
\paragraph{Ablation on multi-task learning:}
Here, we examine whether multi-task learning can improve individual tasks through our unified neural fields. 
We train a single occupancy head, a pose head, and a multi head network.
As shown in Table~\ref{tab:multi_task}, multi-task learning benefits occupancy predictions, suggesting that skeleton prediction may help the occupancy net through the unified neural field representation.

\begin{table}[t]
    \begin{center}
    \begin{tabular}{ccccc}
	\toprule[0.1em]
	\multirow{2}{*}{Head} & \multicolumn{2}{c}{Shape} & & \multicolumn{1}{c}{Pose} \\
	\cmidrule{2-3}\cmidrule{5-5}
    & Chamfer$\downarrow$ & P2S$\downarrow$ & & MPJPE$\downarrow$ \\
    \midrule
    Occupancy & 0.661 & 0.651 & & - \\
    Pose & - & - & & \textbf{1.988} \\
    Occupancy + Pose & \textbf{0.647} & \textbf{0.632} & & 2.051\\
	\bottomrule[0.1em]
    \end{tabular}
    \end{center}
    \vspace{-10pt}
    \caption{Ablation on multi-task learning. Multi-task learning is helpful for shape reconstruction.}
    \label{tab:multi_task}
\end{table}

\cutparagraphup
\paragraph{Ablation on sampling strategy:}
We show the effects of different query points sampling strategy during training for both human surface reconstruction and human pose estimation in Table~\ref{tab:sample_strategy}.
A naive sampling strategy is to uniformly sample the query points (or sample grid points) in the space.
Although this strategy can supervise the neural field in the full 3D space, the query points will be sparse over the human surface or joints location under memory constraints, preventing the network to learn fine-grained surface details and key-points location.
An alternative is to distribute the query points around the object surface (for shape reconstruction) and key-points location (for pose estimation).
We observe that combining the uniform sampling strategy and biased sampling can achieve the best performance for both human surface reconstruction and human pose estimation.
See Figure~\ref{fig:sample_strategy} for a qualitative comparison.
Visually we found that the model trained with uniform sampling alone tends to predict coarse mesh surface and ignore local details.
In contrast, the model trained with biased sampling alone tends to predict artifacts outside the decision boundary, which significantly hurts the Chamfer and P2S results.
The observation is consistent with~\cite{saito2019pifu} where training with hybrid sampling can improve the 3D reconstruction.

\begin{table}[t]
    \begin{center}
    \begin{tabular}{ccccc}
	\toprule[0.1em]
	\multirow{2}{*}{Strategy} & \multicolumn{2}{c}{Shape} & & \multicolumn{1}{c}{Pose} \\
	\cmidrule{2-3}\cmidrule{5-5}
    & Chamfer$\downarrow$ & P2S$\downarrow$ & & MPJPE$\downarrow$ \\
    \midrule
    Uniform & 0.777 & 0.756 & & 3.295 \\
    Biased & 2.503 & 4.340 & & 2.448 \\
    Uniform + Biased & \textbf{0.647} & \textbf{0.632} & & \textbf{2.051} \\
	\bottomrule[0.1em]
    \end{tabular}
    \end{center}
    \vspace{-10pt}
    \caption{Ablation on sampling strategy. The combination of uniform sampling and biased sampling performs best.}
    \label{tab:sample_strategy}
\end{table}

\cutsectionup
\section{Human Animation Details}
\cutsectiondown
\label{sec:animation_details}
We describe the state of the predicted human skeleton as a set of joints $\mathbf J = \{\mathbf j_k\}^K_{k=1}$ and inter-connected bones.
The movement of human skeleton can be represented as a set of relative rotations for the joints in the skeleton.
Given a target pose $\mathbf{\bar{J}} = \{\mathbf{\bar j}_k\}^K_{k=1}$, we first calculate the Inverse Kinematics (IK) that transforms our reconstructed skeleton to this target skeleton.
We denote the solution as $\mathbf \Theta_k \in \text{SO}(3)$, where each relative rotation $\mathbf \Theta_k$ describes how the joint $\mathbf j_k$ will rotate with respect to its parent joint.
Then we animate the human mesh to new pose via the Linear Skinning Model (LBS).
Now we describe our IK solution and the LBS model.

\cutparagraphup
\paragraph{Inverse Kinematics Solution:}
Given the predicted skeleton $\mathbf J = \{\mathbf j_k\}^K_{k=1}$ and target skeleton $\mathbf{\bar{J}} = \{\mathbf{\bar j}_k\}^K_{k=1}$, we aim to find the rotation matrix for each joint that can transform the predicted skeleton to the target skeleton as close as possible.
We solve the inverse kinematics analytically by rotating each joint in the predicted skeleton so that its associated bone (the line connecting the joint and its child joint) is parallel to the corresponding bone in the target skeleton.
Specifically, we use $\mathbf b_k$ to denote the bone direction of the $k$-th joint $\mathbf j_k$ in the predicted skeleton, and $\mathbf{\bar b}_k$ to denote the bone direction of the $k$-th joint $\mathbf{\bar j}_k$ in the target skeleton.
Then the solution $\{\mathbf \Theta_k\}^K_{k=1}$ can be found by solving:
\begin{align}
    \label{eq::roation}
    \mathbf{\bar b}_k = \prod_{p \in A(k)} \mathbf \Theta_p \mathbf b_k, \quad k=1,2,\cdots,K
\end{align}
where $A(k)$ is the set of joint ancestors of the $k$-th joint in order.
It is noted that there are infinitely many rotations that map $\mathbf b_k$ to $\mathbf{\bar b}_k$. We choose the "shortest-arc" rotation between $\mathbf b_k$ and $\mathbf{\bar b}_k$, \ie, rotate along the cross product of $\mathbf b_k$ and $\mathbf{\bar b}_k$.

\cutparagraphup
\paragraph{LBS model:}
Now we describe how we animate the human mesh to new pose using the LBS model.
Formally, we denote the predicted human mesh as a set of $N$ vertices $\mathbf V = \{\mathbf v_i\}^N_{i=1}$, the predicted skeleton as $K$ joints $\mathbf J = \{\mathbf j_k\}^K_{k=1}$, and the predicted skinning weight as a matrix $\mathbf W \in \mathbb R^{N\times K}$.
We then traverse the kinematic tree and construct the rigid transformation matrix $\mathbf T_k(\mathbf \Theta_k)$ for each joint using forward kinematics:
\begin{align}
    \label{eq::chain}
    \mathbf T_k(\mathbf\Theta_k) &= \prod_{p \in A(k)}
    \begin{bmatrix}
    \mathbf \Theta_p & \mathbf (\mathbf I - \mathbf \Theta_p) \mathbf j_p \\
    \mathbf 0 & 1
    \end{bmatrix}
\end{align}
where $A(k)$ is the set of  joint ancestors of the $k$-th joint in order, $\mathbf \Theta_p$ is the rotation matrix of the $p$-th joint \wrt its parent, and $\mathbf j_p$ is the coordinate of the $p$-th joint in the predicted skeleton.
The LBS model assumes the transformation for each vertex $\mathbf v_i$ in the human mesh as a linear combination of the rigid transformation matrix $\mathbf T_k(\mathbf \Theta_k)$ and skinning weight $w_{i,k}$. 
The coordinate for the $i$-th vertex after transformation can now be computed as:
\begin{align}
	\mathbf{\bar{v}}_i=\sum_{k=1}^K w_{i,k} \mathbf T_k(\mathbf\Theta_k) \mathbf v_i
\end{align}
where $w_{i,j}$ is the skinning weight describing the influence of the $k$-th joint on the $i$-th vertex in the predicted mesh. 

\cutsectionup
\section{Additional Results}
\cutsectiondown
\label{sec:additional_results}
We provide more results of our model on real data captured in urban scene with different viewpoints, LiDAR sparsity, lighting and clothes topology. 
The shape reconstruction results are visualized in Figure~\ref{fig:addition_recon}, in each cell from left to right: camera image overlaid with LiDAR points, foreground image, shape reconstruction.
The animation results are visualized in Figure~\ref{fig:addition_anim}, we sample frames from Mixamo~\cite{mixamo} animation as the target skeleton pose, in each cell from left to right: foreground image, reconstructed skeleton, reconstructed shape, target skeleton, re-animated shape.

\end{document}